# Multi-agent coordination for on-demand data gathering with periodic information upload

Yaroslav Marchukov and Luis Montano

Instituto de Investigación en Ingeniería de Aragon (I3A), University of Zaragoza,
C/Mariano Esquillor, s/n, 50018, Zaragoza, Spain
{yamar, montano}@unizar.es

**Abstract.** In this paper we develop a method for planning and coordinating a multi-agent team deployment to periodically gather information on demand. A static operation center (OC) periodically requests information from changing goal locations. The objective is to gather data in the goals and to deliver it to the OC, balancing the refreshing time and the total number of information packages. The system automatically splits the team in two roles: workers to gather data, or collectors to retransmit the data to the OC. The proposed three step method: 1) finds out the best area partition for the workers; 2) obtains the best balance between workers and collectors, and with whom the workers must to communicate, a collector or the OC; 3) computes the best tour for the workers to visit the goals and deliver them to the OC or to a collector in movement. The method is tested in simulations in different scenarios, providing the best area partition algorithm and the best balance between collectors and workers.

## 1 Introduction

In the present work we develop a method for planning and execute a multi-agent team deployment to gather data in some scenario. In every cycle of the the mission the robots have to reach different locations of interest, periodically requested by a static Operation Center (OC), taking measurements and uploading the information to the OC. It will select new goals for the next cycle of request. The OC and the robots have a limited communication range, thus the robots would have to approach to the OC in order to upload the information. A first basic solution is the classical robot deployment towards the objectives, returning all of them to the OC with the new information captured. But for large scenarios it could be very inefficient due to the long time devoted for travelling. An alternative solution is to use some of the robots as information *collectors*, improving this way the time of mission because long journeys are avoided to the rest of robots, which we name *workers*. Depending on times for transmission, for working, and for travelling, a balance between the number of workers and collectors has to be found to minimize the information refreshing time while maximize the number of information packages delivered to the OC. To find a solution to this problem is the objective of the work developed here.

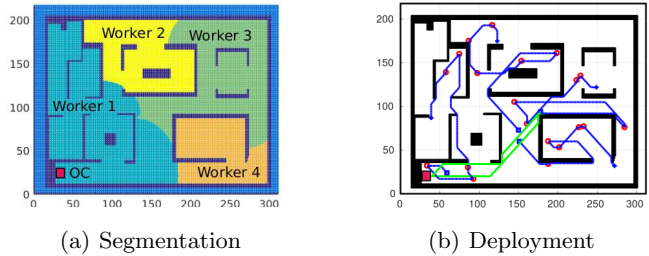

(a) Segmentation  (b) Deployment

**Fig. 1.** Data gathering of 20 objectives with 5 agents: 4 workers and 1 collector, 5 objectives/worker. In (a), the scenario is divided into 4 segments, 1 per worker. (b) depicts the trajectories of 4 workers (blue lines) visiting 5 objectives each (red circles) and going to transmit data. The collector trajectory is the green line.

An assumption is that the goals will be every cycle uniformly distributed within the whole scenario, but changing from one cycle to another. We propose to divide the scenario into working zones for the agents, associating one per worker. This way each worker will receive a number of goals approximately proportional to the size of the area. So the workload of the workers will be also proportional to that size. Although a first idea could be to segment the scenario in zones of similar area to balance the workload of the workers, it does not lead always to the best solution, as we will see later. The segments do not change during the mission, and the workers gather data from the goals of their associated segments, delivering the information either directly to the OC or to the closest moving collector, Fig.1. This way, the team avoids meeting in static rendezvous points to redistribute the working areas of the agents. This would cause a deterioration of the refreshing time of the information. The collectors would need to wait for the workers at each cycle in order to communicate them the distribution of the new segments as well as the next meeting point.

An example of these kind of missions is a fire monitoring, where the locations to monitor and also the frequency of their appearance can change over time. The proposed method can also fit with light changes in other applications such as: human-robot cooperation, where the workers are human operators performing some kind of task in some area and interact during a specified time with a moving robot; warehouse commissioning and logistics, in which the worker robots pickup items and bring them to some collector agents that deliver all the items to the depot point.

The contribution of this work is a technique for planning and executing data gathering missions in connectivity constrained scenarios, working in three steps: (i) the scenario partition in several working areas (so called as segments), each assigned to one worker agent; (ii) the computation of the number of collectors needed for the mission for that partition, their trajectories, and the assignment of workers to meet the collectors or directly to move towards the OC; (iii) the routing for each worker to visit the goals of its assigned area, and synchronize with its assigned collector in movement for the data exchange. The planner is centralized at the OC, but its execution is distributed among the workers for covering its working area and synchronize with the collectors.

For reaching the plan that estimates the best balance between the information refreshing time and the total number of information packages delivered at the OC, the planner evaluates different scenario configurations: three area partition criteria, and different ratios of $\#collectors/\#workers$ in the team. The scenario configuration that provides the best balance is the one selected to be executed.

## 2 Related Work

We can distinguish between two types of communications in multi-agent coordination: permanent or intermittent. The permanent one is usually used in critical missions, such as monitoring in emergency scenarios [12] or formation control for person guidance [10]. Intermittent communication is more usual in exploration missions [1][8], patrolling [3][9], and data gathering[5]. In this work we use intermittent communication due to the size of the scenario, the communication range and the number of agents.

In patrolling missions[3][9], the agents travel invariant paths through a pre-computed graph, re-transmitting the data when they meet each other. Obviously, it will be inefficient to employ this approach for our problem, since the goals appear in different locations, needing to compute the graph with every request cycle. Furthermore, the refreshing time in the OC exponentially increases with each data re-transmission between agents. In our approach, we compute the optimal destinations for the collectors, so they will persistently travel to and from these points and the workers will come to these positions to upload their data, only in the case of having some data to share. So that, only one retransmission is made. This approach is similar in spirit to [6], where an agent performs a persistent task, moving towards other agents to meet them for recharging at the best computed point of their trajectories. In our kind of missions, it is more effective that the collector agents travel invariant paths, being the workers who move to share the data with them, in order to preserve the time of collectors cycles since it delimits the refreshing time.

In [5], the authors develop a method where some workers, with buffer limitations, gather data and transmit it to dynamic collectors. Their collectors are permanently connected to a central server in the entire scenario. They are the only ones capable to upload the data to the server without the need of travel to a unique static depot point, situation considered in our work. Obviously, the travels to a static depot point increase the refreshing time, especially in large scenarios. In our approach the workers remain in their working areas and only travel short paths to the collectors, not needing to go up the OC.

Our method is more flexible because the workers are not enforced to concur with its collector to some fixed rendezvous points, as in classic agent meeting problem[7], since it may become inefficient for big fleets of agents. Instead, the workers decide when and where to share the data with the collector, so that not stopping the motion if it is not needed. In [2], the robots are enabled to transfer deliveries between each other to efficiently reach different delivery locations. However, their system to establish the meeting points is fully centralized, that cannot be directly applied here because our team acts in a distributed way during the execution of the mission.

## 3 Problem Statement

### 3.1 Problem Formulation

The problem to solve is planning the deployment of a team of robots in a scenario with connectivity constraints due to obstacles or distances to the OC. This requests for data from different goals to be periodically uploaded them in the OC. We use a grid representation for the scenarios because, as justified bellow, we apply the Fast Marching Method (FMM)[11] as a common tool for several algorithms of our method. We denote a position in the grid as $x$ and a set of positions as $\mathbf{x}$. The positions of the obstacles are denoted as $\mathbf{x}_o$, the position of the OC is expressed as $x_{oc}$, and the positions of the agents and goals are $\mathbf{x}_a$ and $\mathbf{x}_g$, respectively. The OC periodically requests information from $M$ goal locations, and the team, composed by $N$ robots, is coordinated to move towards the goals, then delivering the information to the OC. The robots can act either as workers ($N_w$) or as the collectors ($N_c$), being $N_c + N_w = N$. During the time of the mission, denoted as $T_{mission}$, $M$ remains constant. That is, when OC receives the information from $m$ goals ($m \leq M$), it generates $m$ new goals. Our approach must compute the plan of the mission, previously to deploy the agents. To this end, it must: i) obtain the working areas for the worker agents, denoted as $S_w$; ii) find out the best balance between collector and workers; iii) pair the workers with the collectors or with the OC to transmit the data, expressed as $P_{cw}$; iv) compute the trajectories of each collector $\pi_c$, according to $P_{cw}$ and $S_w$. Once the mission starts, each worker computes its path $\pi_w$, in order to visit the corresponding goals, $\mathbf{x}_{g_i} \in S_{w_i}$ and go to deliver the information to a collector or OC, according to the association $P_{cw}$.

$\Pi_w$ and $\Pi_c$ are the sets of worker's and collector's paths, respectively, being $\pi_w \in \Pi_w$ and $\pi_c \in \Pi_c$. The times for travelling a path $\pi$ and paths $\Pi$ are expressed as $t(\pi)$ and $t(\Pi)$, respectively. The refreshing period from the time in which OC requests information and receives it is $T_{refresh}$. This period is the mean time of the collectors $T_c$ and of the workers that transmit directly to the OC. Naturally, it must be minimal, whilst the number $m$ of goal information received has to be as large as possible. The problem of computation of trajectories of workers and the collectors can be formally expressed as:

$$\pi^*_{w_{ij}} = \underset{\pi_{w_i} \in \Pi_{w_i},\ \mathbf{x}_{g_i} \in \pi_{w_i}}{\operatorname{argmax}} \left( |\mathbf{x}_{g_i}| \right) \qquad (1)$$

$$\pi^*_{c_j} = \underset{\pi_{c_j} \in \Pi_{c_j}}{\operatorname{argmin}} \left( t(\pi_{c_j}) - \overline{t(\Pi_{w_{ij}})} \right) \qquad (2)$$

$$subject\ to\ t(\pi^*_{c_j}) - t(\pi^*_{w_{ij}}) \geq 0 \qquad (3)$$

where $c_j$ refers to collector $j$ and $w_i$ to worker $i$. Eq.(1) obtains the optimal route for workers to visit the maximum number of goals $\mathbf{x}_{g_i}$ assigned to the agent

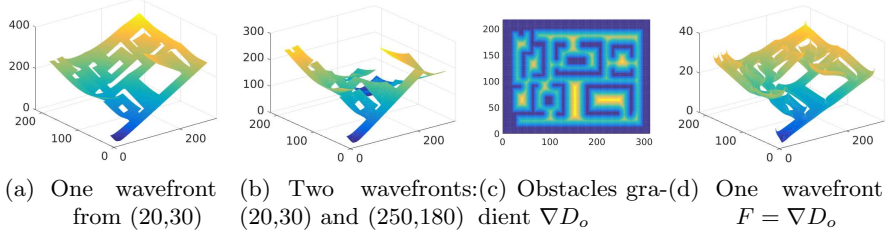

(a) One wavefront from (20,30)  (b) Two wavefronts: (20,30) and (250,180)  (c) Obstacles gradient $\nabla D_o$  (d) One wavefront $F = \nabla D_o$

**Fig. 2.** FMM gradients.

$w_i$ from all the possibles $\Pi_{w_i}$. Eq.(2), computes the trajectory of the collector $c_j$ that minimizes the time the workers must wait for the collector arrival, once that they have visited the goals within its area. $\Pi_{w_{ij}}$ represents the paths of the workers $w_i$ assigned to collector $c_j$, obtained with eq.(1), and $\overline{t(\Pi_{w_{ij}})}$ denotes the mean time of these paths. Eq.(3) constraint enforces the workers to fulfill the maximum number of their assigned goals, eq.(1), to meet the collector when it arrives in the current cycle. When a worker and a collector meet each other, the distance between them must be lower than $d_{com}$ and the line-of-sight must not be occluded to establish communication. The three steps achieved by the planner are described in the following sections: i) scenario segmentation in Sect.4, ii) collector's trajectories and segment allocation to workers in Sect.5, and iii) workers routing in Sect.6.

### 3.2 Fast Marching Method (FMM)

The FMM is applied for several planner steps: the area partition, the path planning for the collectors, and the computation of the best route to the workers visit the goals and synchronize for data delivery. The basic idea of FMM is the propagation of a wavefront from some position $x$ over a static grid, computing a distance gradient $\nabla D$:

$$|\nabla D(x)|F = 1 \qquad (4)$$

The wavefront is propagated over every point of the grid with a velocity $F$, solving eq.(4). In a simple grid, $F$ takes values of: 0 in positions with an obstacle, and 1 in free space. Hence the wavefront is propagated uniformly in all the directions, surrounding the obstacles, obtaining the distance gradient $\nabla D$, Fig.2(a). Descending this gradient from some goal position $x_{goal}$, we obtain a path from the origin to $x_{goal}$.

Initializing several wavefronts, the resulting $\nabla D$ will represent the distance to the closest origin, see Fig.2(b). This is used for the segmentation methods of Sect.4.2-4.3. The positions where the wavefronts collide, are the frontiers of the segments. Varying the values of $F$, the propagation velocity is non-uniform. The gradient propagates faster for higher values and slower for lower values of $F$. An example is depicted in Fig.2(c)-2(d). The obstacles gradient $\nabla D_o$ is obtained with FMM, Fig.2(c). Propagating the wavefront, with $F = \nabla D_o$, the resultant gradient obtains lower values in points that are more distant from the obstacles,

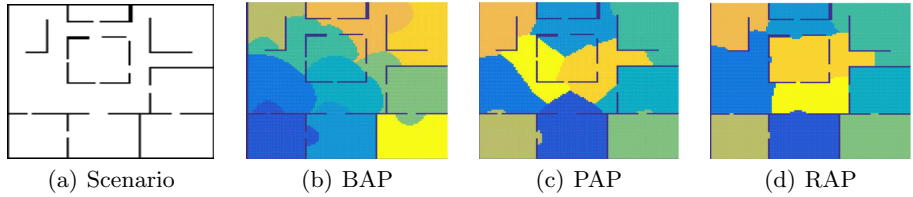

**Fig. 3.** Segmentation algorithms for 10 segments.

Fig.2(d). This property is also used for the segmentation and for associations between collectors and workers in Sect.5.3.

## 4 Scenario Segmentation

In this work we develop and evaluate three segmentation algorithms: $BAP$, $PAP$ and $RAP$. All of them use the base method FMM. We encourage to visit the link [1], with the proposed segmentation processes in different scenarios.

### 4.1 Balanced Area Partition (BAP)

The main feature of this segmentation algorithm is that it obtains segments with balanced areas. A uniform FMM wavefront propagation obtains these areas. Firstly, the algorithm computes the gradient from the OC position $\nabla D_{OC}$, which denotes the distance to the OC. Since the team employs $N_w$ workers, the space is divided into $N_w$ segments, with areas $A_{w_i}, i = 1, ..., N_w$, accomplishing $A = \sum A_{w_i}$. The optimal segment area is denoted by $A_{opt} = A/N_w$, measured as number of cells in the grid.

Initially FMM propagates a wavefront from OC, expanding $A_{opt}$ cells. If the total area of expanded cells is higher or equal to $A_{opt}/2$, a heuristic threshold, and the number of already obtained segments is lower than $N_w$, these cells become a new segment. If not, the expanded cells are added to an adjacent segment of minimum area.

The algorithm iterates until classify all the free space $A$, choosing as wavefront origin, the closest non-classified point to $x_{OC}$ from $\nabla D_{OC}$. The number of extended cells depends on the obstacle distribution in the scenario, since the algorithm requires more iterations to cover the entire area if there are any remaining non-classified small areas. This may also produce bigger segments than $A_{opt}$, obtaining a lower number of segments than $N_w$. In this case, the algorithm iteratively halves the biggest segments until obtain $N_w$. Fig.3(b) illustrates the segmentation for 10 segments. Although the shape of the segments is quite irregular, the areas are equitable, which a priori favours a balanced time for workers in all the segments.

### 4.2 Polygonal Area Partition (PAP)

---

[1] http://robots.unizar.es/data/videos/paams19yamar/segmentations/

**Algorithm 1** Procedure for PAP and RAP

**Require:** Grid, $\mathbf{x}_o$, $N_w$, Partition type ($PAP$ or $RAP$)
1: $\nabla D_o \leftarrow compute\_gradient(\mathbf{x}_o)$ ▷ Eq.(4)
2: $\mathbf{x}_c \leftarrow initialize\_centroids(N_w, \nabla D_o)$
3: **If** ($PAP$) $[S, \mathbf{x}_c] \leftarrow it\_part(\mathbf{x}_c, Grid)$ ▷ Alg.2
4: **If** ($RAP$) $[S, \mathbf{x}_c] \leftarrow it\_part(\mathbf{x}_c, \nabla D_o)$ ▷ Alg.2
5: **return** $S, \mathbf{x}_c$

This algorithm attempts to keep equitable distances between the centroids of the segments and their boundaries with other segments or obstacles. $PAP$ algorithm is summarized in Alg.1. The algorithm consists in two main phases: centroid initialization and iteration phase. Firstly, the algorithm obtains the obstacles gradient, corresponding to the distance to them, l.1, as explained in Sect.3.2, see Fig.2(c). After that, it iteratively finds the maximum values of this gradient, l.2, which correspond to the largest free spaces, for example rooms. The position of the maximum is the initial location of the centroid of each segment. Since the value of the maximum represents the greatest distance to the closest obstacle, the algorithm removes all the points of the grid within the radius equal to the maximum. This way, we avoid to initialize new centroids of the segments in the same rooms. This procedure is repeated $N_w$ times, one per segment/worker.

After the initialization, the method iteratively moves the centroids until achieving the equilibrium between the distances of the centroids. This procedure is described in Alg.2. At first, the method computes the distance gradient from the centroids (l.2), using again FMM. The variable *costmap* takes the values of the basic grid to propagate the wavefronts, so $F$ is 0 or 1 in eq.(4). Therefore, $N_w$ wavefronts are uniformly propagated in all the directions, one per segment. The positions where the wavefronts collide among them are the boundaries of the segments. The highest value of the gradient of each partition is the farthest position from the centroid $\mathbf{x}_{p_f}$ in l.3. So we move every centroid in the direction of its farthest position in the segment: computing the paths, l.4, and moving the centroids along them, l.5. We encourage the reader to watch the video for better understanding of this procedure. The resulting segments are depicted in Fig.3(c).

### 4.3 Room-like Area Partition (RAP)

**Algorithm 2** Iterative Partition (*it_part*)

**Require:** $Centroids\ (\mathbf{x}_c), costmap$
1: **while** $!repeated\_positions(\mathbf{x}_c)$ **do**
2:   $[\nabla D_c, S] \leftarrow gradient\_and\_partitions(\mathbf{x}_c, costmap)$
3:   $\mathbf{x}_{pf} \leftarrow compute\_farthest\_in\_partition(\nabla D_c, S)$
4:   $\Pi_f \leftarrow gradient\_descent(\nabla D_c, \mathbf{x}_{pf})$
5:   $\mathbf{x}_c \leftarrow move(\Pi_f)$
6: **end while**
7: **return** $S, \mathbf{x}_c$

This area partition method employs the same procedure that iteratively moves the centroids as $PAP$. But instead of using the basic grid of the map, in l.4 of Alg.1, it uses the obstacles gradient $\nabla D_o$ computed in l.1. Then, *costmap* variable takes values of $\nabla D_o$ in Alg.2. This changes the propagation of the wavefronts, becoming non-uniform. So that the wavefronts cover faster the wide areas, such as rooms, and slow down when reach tight spaces, commonly corresponding to doors, where the wavefronts collide. Using this property, the resulting segments tend to cover the rooms, as can be seen in Fig.3(d). Because of that, their areas differ from $PAP$'s algorithm.

## 5 Collectors trajectories and segment allocation

In this section we explain how the planner selects the best number of the collectors used for the mission (based on the workers segments), the computation of their trajectories, and the association of the workers to the collectors to share data.

### 5.1 Working time estimation

The method computes the collectors trajectories based on an estimated working time of their associated workers. During the mission the collectors permanently travel these paths without stopping, being the workers who move to share the gathered data with them.

Since the distribution of the goals will change every cycle, the planner has to estimate an averaged working time in each segment. Assuming the goals are uniformly distributed, we consider the number of goals within each worker segment $i$ will be approximately proportional to its area, obtained as $M_{s_i} = M * A_{s_i}/A$. In order to estimate a fair approximation of the distribution of the goals within a segment, the algorithm automatically places $M_{s_i}$ goals (centroids) within segment $S_i$ using the $PAP$ procedure. This way, the estimated goals are equidistant between them, considering the obstacles. The $NN+2O$ procedure, explained in Sect.6.1, computes the shortest tour from each segment centroid to visit the $M_{s_i}$ goals, estimating the working time each worker will spend in its segment.

### 5.2 Planning procedure

**Algorithm 3** General planning procedure

**Require:** $Grid$, $M$, $N$, $x_{oc}$
1: $\Pi_c^* \leftarrow \emptyset$, $P_{cw}^* \leftarrow \emptyset$, $S_w^* \leftarrow \emptyset$, $\mathbb{T}_c \leftarrow \emptyset$, $\mathbb{M}_d \leftarrow \emptyset$
2: $\nabla D_{oc} \leftarrow compute\_gradient(x_{oc}, Grid)$
3: **for each** $segmentation$ **do** ▷ BAP, PAP, RAP
4:    **for** $N_c = 0 : N/2$ **do**
5:       $N_w = N - N_c$
6:       $[S_w, \mathbf{x}_{c_w}] \leftarrow segment(grid, N_w)$
7:       $T_{work} \leftarrow estim\_work\_time(S_w)$ ▷ Sect.5.1
8:       $[\Pi_c, P_{cw}, T_c, M_d] \leftarrow coll\_paths(Grid, N_c,$
          $...x_{oc}, \nabla D_{oc}, \mathbf{x}_{c_w}, S_w, T_{work})$
9:       $\mathbb{T}_c \leftarrow \mathbb{T}_c \cup \{T_c\}, \mathbb{M}_d \leftarrow \mathbb{M}_d \cup \{M_d\}$
10:    **end for**
11: **end for**
12: $[\Pi_c^*, S_w^*, P_{cw}^*] \leftarrow best\_plan$ ▷ eq.(5)
13: **return** $\Pi_c^*, S_w^*, P_{cw}^*$

Based on the average working time estimated for the segments, the plan procedure in Alg.3 obtains the needed collectors, their time periods $T_c$, and their trajectories. The paths of collectors are computed from the gradient to the OC, l.2. It computes the three scenario partitions, l.3. The number of collectors to be evaluated is $N_c = 0..N/2$, l.4, because it makes no sense to devote more than one collector to a single segment. When the system adds a new collector, it renounces to a worker,

l.5. This changes the working areas, so the algorithm segments the scenario every iteration, l.6, estimating the working times for the resulting segments, l.7 (Sect.5.1). Then it computes the paths of the collectors ($\Pi_c$) and collector times ($T_c$), l.8, and associates the workers with the collectors ($P_{cw}$), obtaining the goals that will be delivered to the OC ($M_d$), explained in Sect.5.3. A direct movement of workers to the OC, without using a collector, is also evaluated ($N_c = 0$). The times and the delivered goals are stored, l.9, in order to choose the best plan in l.12 from the different partitions and collectors, using the utility function:

$$U = max\big[\alpha * (1 - \mathbb{T}_c/max(\mathbb{T}_c)) + \beta * \mathbb{M}_d/max(\mathbb{M}_d)\big], \ \alpha + \beta = 1 \quad (5)$$

In this work, we set the values $\alpha = \beta = 0.5$, giving the same priority to the refreshing time and the number of deliveries. However, these values can be adjusted depending on the kind of mission, i.e. higher values for $\alpha$ in critical missions, such as fire monitoring, and higher values for $\beta$ in simpler missions, such as surveillance.

### 5.3 Collector trajectories and workers-collectors association

An illustrative example of collector computation is depicted in Fig.4. This method also employs a FMM-based $RAP$ procedure to associate the collectors to the workers. It receives the segmented scenario for the workers, Fig.4(a), and obtains the graph of the worker segments, Fig.4(b). The vertices of the graph are the centroids of the segments, the edges joining the adjacent vertices. The work-ers at the segment of the OC and the adjacent ones will upload data directly to the OC, thus their edges are removed from the graph.

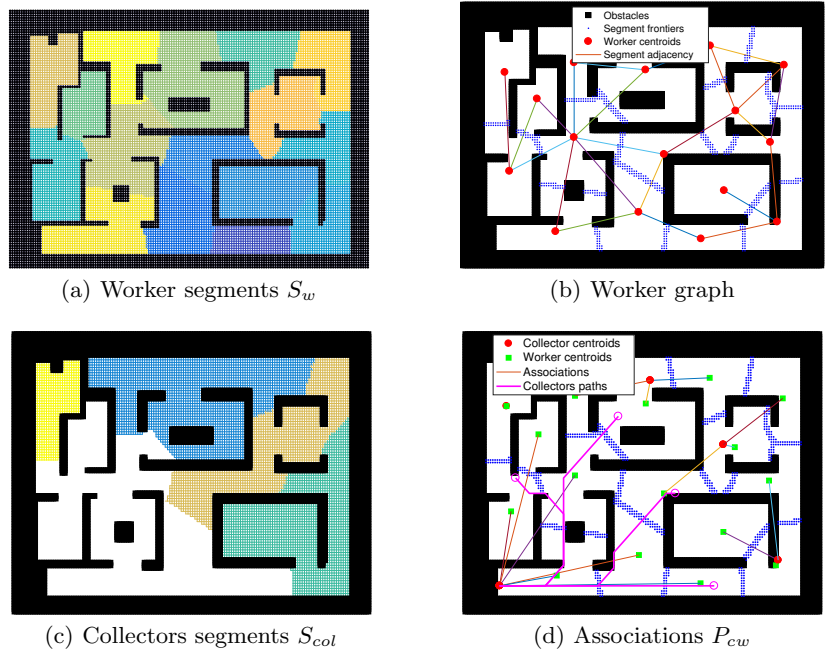

(a) Worker segments $S_w$  
(b) Worker graph  
(c) Collectors segments $S_{col}$  
(d) Associations $P_{cw}$

**Fig. 4.** Collectors' path computation and association, for 16 segments (workers) and 4 collectors. In (c), white space are segments of workers that upload directly to the OC

The method iteratively computes the paths of the collectors, associates them with the workers, and estimates the mean refreshing time at the OC. First, it computes the connectivity of the graph, that is the number of edges that has every vertex. Second, it selects $N_C$ vertices of the maximum connectivity, and initializes the centroid of the segments as the initial farthest position to be reached by the collector, $\mathbf{x}_{col}$, before come back to the OC. Then it iteratively moves these centroids, by propagating the gradient $\nabla D_{front}$ from the frontiers of the worker segments, depicted with blue points in Fig.4(b), using Alg.2 executing $[S_{col}, \mathbf{x}_{col}] \leftarrow it\_part(\mathbf{x}_{col}, \nabla D_{front})$. This way, the shape of the segments is taken into account, filling faster the smaller and uniform worker segments and slower the largest and irregular ones. The final segments associated to the collectors ($S_{col}$), which groups the worker segments associated to it, are depicted in Fig.4(c). The workers of the segments grouped in a collector segment $S_{col}$ share data with the collector that comes to $\mathbf{x}_{col}$, Fig.4(d). Then, the paths from $x_{OC}$ to each $\mathbf{x}_{col}$ are obtained descending $\nabla D_{OC}$. According to eq.(1)-(3), in order to find out a balance between the time of the collector $T_c$ and time of workers to reach $\mathbf{x}_{col}$, each collector path is iteratively contracted until achieving the balance. As can be observed in Fig.4(c), the method associates not only adjacent workers to upload to OC, but also others corresponding to the next level in the graph until $T_{refresh}$ stops decreasing.

## 6 Workers trajectories

Here we develop the trajectory planner that execute the workers to visit the goals and to synchronize with the collector in movement.

### 6.1 Goals visit methods

Each worker has to visit the maximum number of goals in its segment, in such a way that it is able to reach the collector to share the data in some point of its trajectory. The costs to reach the goals are evaluated from the FMM gradient; the distances are the values of the gradient at the goals positions. We use three main routines to obtain the worker's route:

- *Brute Force (BF)*: obtains the optimal solution testing all the possible routes.
- *Nearest Neighbor with 2-opt Improvement (NN+2O)*: a first route is initialized with the *Nearest Neighbor (NN)* procedure. Then, the route is improved by means of a local optimization using 2-opt method [4], that swaps every two edges of the route, goals in our case, checking if the new route outperforms the previous one. This method is able to obtain the routes in real-time (milliseconds), against classic orienteering problem methods, which require minutes to find a solution [13].
- *NN+2O with Time Window (NN+2O-TW), and BF-TW*: with the same structure as the basic *NN+2O* and *BF*, but taking into account a time condition to reach the collector at time, formally expressed as:

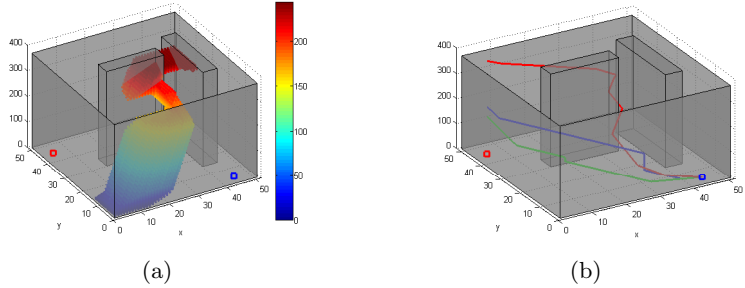

(a)                                (b)

**Fig. 5.** Workers synchronization. In (a), a collector, in movement, creates a dynamic communication area. In (b), a worker goes from blue to red circle, obtaining *Intercepting*, *Waiting*, and *Following* trajectories for $t_{tx} = \{10, 30, 100\}$ sec, are green, blue and red lines, respectively.

$$T_c - t_{tx} >= t_{accum} + t_{g_j} + t_{g_j-c}, j = 1, ..., K \quad (6)$$

where $T_c$ is cycle time for the collector, $t_{tx}$ is the time to transmit the data of already collected information, $t_{accum}$ is the accumulated time of the trajectory, $t_{g_j}$ is the time to reach the next non-visited goal $j$, $t_{g_j-c}$ is the time from this goal to the collector and $K$ is the number of goals for the worker. When the condition of eq.(6) is not accomplished, the algorithm stops iterating.

The workers that upload data directly to OC use *BF* to obtain a solution in real-time for instances up to 12 goals (about 50ms), whilst *NN+2O* is used to obtain a suboptimal solution for more than 12 goals. The workers that transmit to a collector employ *BF-TW* for less than 12 goals, otherwise use *NN+2O-TW*. This way the workers upload to the collector every cycle some or all the tasks allocated for them.

### 6.2 Trajectories for synchronization

To transmit the data, the workers must remain within the collector communication area during $t_{tx}$ while the collectors are moving Fig.5(a). This way the agents do not use static meeting points, but synchronize with the collectors in dynamic rendezvous areas, chosen by the workers at each gathering cycle, based on the amount of gathered data to transmit and on the time to meet the collector in a point along its path. The FMM is used to obtain the optimal trajectories to synchronize with a collector in movement to transmit the data. The workers act in three different ways depending on $t_{tx}$: i) *Intercepting*: if $t_{tx}$ is much lower than the vertical section of the collectors communication area, so, the worker can simply traverse it (green line in Fig.5(b)); ii) *Waiting*: if $t_{tx}$ is equal to the vertical section, the worker can await within the area until it transmits the data (blue line); iii) *Following*: if $t_{tx}$ is higher than the vertical section, and the worker must follow the collector until fulfill the data transmission (red line).

Note, that if there is an uncertainty in the collector's trajectory or even if it fails during its motion, the worker follows the theoretical collector path until finding it, or up to arriving to OC. This way the mission will be achieved, although the refreshing time will increase and the number of delivered goals during the mission will be reduced.

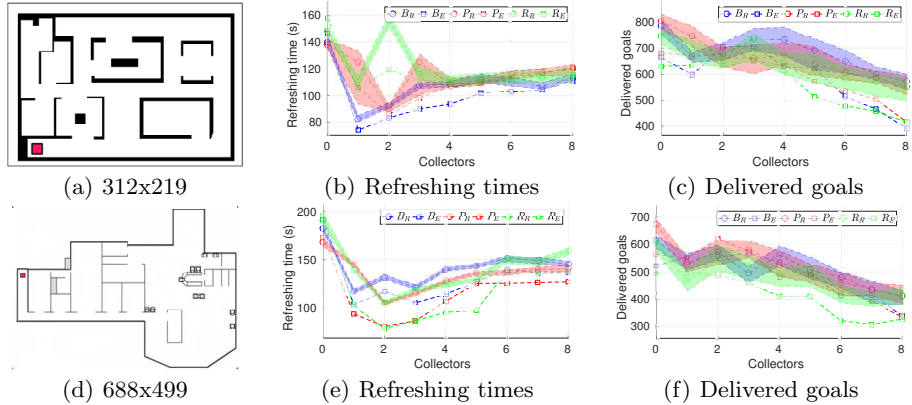

**Fig. 6.** Results. The red square in (a) and (d) is the OC. The letters $B, P, R$ in the legends refer to $BAP, PAP, RAP$ methods respectively. Sub-index $R$ and $E$, denote real and estimated values, respectively. The squares represent estimations and the circles the real values. The coloured bands are delimited by the maximum and minimum of the real values.

## 7 Results

The method was implemented in C++ and performed in a computer with i7 CPU clocked at 3.4GHz with 8GB of RAM. We evaluate it by means of simulations in the scenarios of Fig.6(a) and Fig.6(d) (extracted from [3]). The results for the planner of Alg. 3 are evaluated for $N=20$ agents and $M=100$ requested goals, $d_{com}=10$ cells and a constant velocity for the workers and collectors of 2 cell/sec. The time to gather data at one goal is 5 seconds and to transmit this data is 1 second. The plan execution is also evaluated, based on 20 trials of randomly generated goals, in a 1000 seconds mission. The mission starts with the agents at OC, the plan is obtained with Alg.3 and shared between the agents. Then, in a initialization phase, the workers go to their respective segments and start gathering the first batch of requested goals. The results analysis shows that the mean refreshing times and number of packages delivered at OC for different workers-collectors ratios and the three partition methods, are close to the ones estimated by the planner. Examples of deployments are available in the link [2].

The results are depicted in Fig.6. Both scenarios present different layouts and difficulties for the team deployment. The mean time to obtain the plan with Alg.3, is less than one minute in both scenarios. The complexity of Alg.3 is $O(N/2(b_w + b_c + 1)nlogn)$, where: $n$ is the number of free cells in the map, being $nlogn$ the complexity of FMM; $b_w$ is the number of iterations to achieve the balance in the segmentation process for the workers ($b_w = 1$ for $BAP$); $b_c$ are the iterations to balance the collector segments; the remaining +1 is the FMM to obtain the paths of the collectors. The algorithm tests different number of collectors from 0 (baseline for comparison) to 8 ($N/2$). The complexity to obtain the trajectories for a worker $i$ at each cycle is: $O((N_{g_i} + 3)nlogn)$. Requiring $N_{g_i} + 3$ FMM executions, being $N_{g_i}$ the number of goals of the agent $i$, and the remaining 3 are the FMM's computations: from the agent position, within the communication area, and to the next goal location.

---

[2] http://robots.unizar.es/data/videos/paams19yamar/simulations/

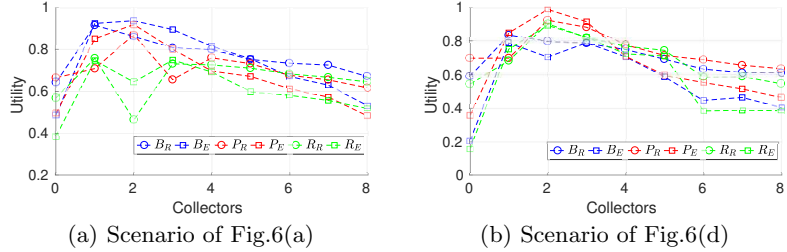

**Fig. 7.** Utilities, using eq.(5)

According to the utilities of Fig.7 (eq.(5)), a first clear result is that using collectors is better than not using them. The planner estimates that the best plan for the scenario of Fig.6(a) is $BAP$ using 2 collectors, although 1 or 3 collectors provide similar utilities. For Fig.6(d) the best is $PAP$ with 2 collectors. The computed utilities for the plan execution are close to the planned ones: the best result in Fig.6(a) is $BAP$ with 1 collector, and for Fig.6(d) the execution values match with the estimated ones. Regarding Fig.6, in mean the plan execution get worse $T_{refresh}$, the time elapsed to deliver the requested the information, only in 9 seconds, delivering 10 less goals. The little differences found between planned and execution mean values are due to the estimation considers that the number of goals within each segment is proportional to its area, that does not always occur with real goals distribution. It can be concluded that the use of 1-3 collectors provide the best results in the tested scenarios.

Regarding the kind of segmentation, the $BAP$ method that splits the scenario in segments of approximately similar area, works better in the first scenario in which a more homogeneous obstacle distribution is found. The $PAP$ method, which splits the scenario in polygonal segments, works better in the second scenario, where the obstacles are not homogeneously distributed, having large diaphanous areas and narrow corridors. The polygonal segmentation fits better the cleared areas than the other segmentation methods. Anyway, seeing the refreshing time, the number of goals delivered, and the utilities, it can be said that the $PAP$ segmentation provides a good solution for both scenarios.

## 8 Conclusions

In this paper we have presented a method to plan the deployment of a team of agents to periodically gather information on demand from some a priori unknown goal locations, delivering them to a static operation center. The use of collectors for uploading the information at OC is more useful that directly moving all the robots to the OC, from the point of view of the balance between the refreshing time and the number of delivered goals. We have tested three area partition algorithms, concluding that the $PAP$ segmentation, which splits the scenarios in polygonal areas that fit well the free workspaces, provides good results for one to three collectors in the tested scenarios. As future works, we will include the uncertainty of the trajectories of the collectors for the synchronization with the workers, and will use a training phase to estimate from examples the goals distribution in the scenario.

## Acknowledgement

This research has been funded by project DPI2016-76676-R-AEI/FEDER-UE and by research grant BES-2013-067405 of MINECO-FEDER, and by project Grupo DGA-T45-17R/FSE